\documentclass{article}\usepackage[]{graphicx}\usepackage[]{color}
\makeatletter
\def\maxwidth{ %
  \ifdim\Gin@nat@width>\linewidth
    \linewidth
  \else
    \Gin@nat@width
  \fi
}
\makeatother

\definecolor{fgcolor}{rgb}{0.345, 0.345, 0.345}

\usepackage{framed}
\makeatletter
 {\par\unskip\endMakeFramed%
 \at@end@of@kframe}
\makeatother

\definecolor{shadecolor}{rgb}{.97, .97, .97}
\definecolor{messagecolor}{rgb}{0, 0, 0}
\definecolor{warningcolor}{rgb}{1, 0, 1}
\definecolor{errorcolor}{rgb}{1, 0, 0}
\newenvironment{knitrout}{}{} 

\usepackage{alltt} 

\usepackage[authoryear]{natbib}
\usepackage{times}

\usepackage{hyperref}
\usepackage{url}

\usepackage[margin=1.5in]{geometry}

\usepackage{times}
\usepackage{graphicx} 
\usepackage{subcaption}
\usepackage{caption}

\usepackage{amsmath}
\usepackage{amsthm}
\usepackage{appendix}

\usepackage{amssymb}
\usepackage{bbm}

\usepackage[accepted]{data4good2016}
\newcommand{\mysec}[1]{Section~\ref{sec:#1}}

\newcommand{\eq}[1]{Eq.~(\ref{eq:#1})}

\newcommand{\fig}[1]{Fig.~(\ref{fig:#1})}

\newcommand{\mbe}{\mathbb{E}}
\newcommand{\mbeq}{\mathbb{E}_{q}}

\newcommand{\iid}{\stackrel{iid}{\sim}}
\newcommand{\indep}{\stackrel{indep}{\sim}}

\newcommand{\pthetapost}[1][\alpha]{p_{x}^{#1}}
\newcommand{\qthetapost}[1][\alpha]{q_{x}^{#1}}
\newcommand{\gtheta}{g\left(\theta\right)}
\newcommand{\expectp}[2][\alpha]{\mathbb{E}_{\pthetapost[#1]}\left[#2\right]}
\newcommand{\expectq}[2][\alpha]{\mathbb{E}_{\qthetapost[#1]}\left[#2\right]}
\newcommand{\epgtheta}[1][\alpha]{\expectp[#1]{\gtheta}}

\theoremstyle{plain}

\IfFileExists{upquote.sty}{\usepackage{upquote}}{}
\begin{document}

\twocolumn[
\icmltitle{Fast robustness quantification with variational Bayes}

\icmlauthor{Ryan Giordano}{rgiordano@berkeley.edu}
\icmladdress{UC Berkeley}
\icmlauthor{Tamara Broderick}{tbroderick@csail.mit.edu}
\icmladdress{MIT}
\icmlauthor{Rachael Meager}{rmeager@mit.edu}
\icmladdress{MIT}
\icmlauthor{Jonathan Huggins}{jhuggins@mit.edu}
\icmladdress{MIT}
\icmlauthor{Michael Jordan}{jordan@cs.berkeley.edu}
\icmladdress{UC Berkeley}

\icmlkeywords{boring formatting information, machine learning, ICML}

\vskip 0.3in
]

\newcommand{\fix}{\marginpar{FIX}}
\newcommand{\new}{\marginpar{NEW}}



\begin{abstract}
Bayesian hierarchical models are increasing popular in economics.
When using hierarchical models, it is useful not only to calculate posterior
expectations, but
also to measure the robustness of these expectations to reasonable alternative
prior choices.  We use variational Bayes and linear response methods to
provide fast, accurate posterior means and robustness measures with
an application to measuring the effectiveness of microcredit in the developing world.
\end{abstract}

\section{Introduction}\label{sec:intro}

Researchers and policymakers in economics increasingly have access to results
from several experimental studies of the same phenomenon. Particularly in
development economics, following the recent proliferation of randomized
controlled trials to study key anti-poverty interventions, the question of how
to aggregate the results of multiple experiments across different contexts has
now arisen. Recent attempts to perform this aggregation have noted that
different studies of the same intervention often produce different results, but
both the extent of the true variation in the underlying treatment effects and
the source of such variation are often unclear \citep{meager:2015:microcredit,vivalt:2015:economics,burke:2015:climate}. There is however a methodology which is ideally suited to aggregating evidence
and assessing the extent of heterogeneity across contexts, and has been well
developed by statisticians: Bayesian hierarchical models
\citep{rubin:1981:estimation,gelman:1992:inference}.  Bayesian analysis turns
the data model likelihood and a distribution of prior beliefs into a posterior
distribution over the model parameters through application of Bayes' rule.
Often, the posterior is summarized by certain moments of the model parameters
(e.g. the mean or variance).

However, with moderately large datasets, Bayesian hierarchical model posteriors
can be time-consuming to estimate even using cutting-edge Markov Chain Monte
Carlo (MCMC) software such as Stan \cite{stan-manual:2015}.  Futhermore, lower
levels of the hierarchical model, which often represent the quantities of
practical interest, can be sensitive to the choice of priors, leading to
\emph{non-robust posteriors}.  If different reasonable choices of the prior lead
to substantially different posterior means, then the model is not robust, since
different arbitrary prior choices could lead to different substantive
conclusions. It is important for the modeler to be aware of non-robustness,
either so that the model can be improved or the conclusions qualified. Though
procedures exist to measure robustness with MCMC draws, an easy-to-use,
general-purpose methodology is still lacking \citep{berger:2012:robust}.

In this paper we take a step towards addressing the slow estimation times of
hierarchical models and provide automated measures of robustness using
\emph{variational Bayes} (VB). VB is an optimization-based method for performing
approximate Bayesian posterior inference \citep{wainwright2008graphical,
bishop:2006:pattern}. In contrast to MCMC, which produces draws from a
distribution that approaches the true posterior asymptotically, VB finds the
best approximation to the true posterior within a restricted class of
distributions.  Since VB is an optimization procedure rather than a Markov
chain, it can often produce posterior approximations much more quickly than
MCMC, though at the cost of providing only an approximation, even
asymptotically.

In addition to being generally faster than MCMC, variational Bayes (VB)
techniques are more readily amenable to robustness analysis.  The derivative of
a posterior expectation with respect to a perturbation of the prior or the data
is a measure of \emph{local robustness} to the prior or likelihood
\citep{gustafson:2012:localrobustnessbook}. Because VB casts posterior inference
as an optimization problem, its methodology is built on the ability to calculate
derivatives of posterior quantities with respect to model parameters, even in
very complex models.

In order to provide fast estimates of local robustness, we use the machinery of
\emph{Linear response variational Bayes} (LRVB), previously developed by
\citep{giordano:2015:lrvb}. \citet{giordano:2015:lrvb} show how perturbing the
posterior distribution can yield improved estimates of posterior covariance over
vanilla VB. In this work, we demonstrate that a similar idea can be applied to
derive fast, easy-to-use robustness measures.

In the remainder of this work, we start by briefly describing VB and LRVB in
\mysec{basic_lrvb}.  In \mysec{robustness} we describe how to measure local
robustness using LRVB.  Finally, in \mysec{microcredit}, we demonstrate our
methods on a meta-analysis of  data from seven randomized controlled trials of
microcredit expansions.

\section{Variational Bayes and linear response}\label{sec:basic_lrvb}

Denote our $N$ data points by $x = (x_1, \ldots, x_N)$ with $x_n \in
\mathbb{R}^{D}$. Denote our parameter by the vector $\theta \in \mathbb{R}^{K}$.
We denote the prior parameters by $\alpha \in \mathbb{R}^{M}$. Let $\pthetapost$
denote the posterior distribution of $\theta$, as given by Bayes' Theorem:
\begin{eqnarray*}
  \pthetapost\left(\theta\right) :=
  p\left(\theta \vert x, \alpha \right) =
    \frac{p\left(x \vert \theta \right) p\left(\theta \vert \alpha \right)}
    {p\left(x\right)}.
\end{eqnarray*}
VB approximates $\pthetapost\left(\theta\right)$ by selecting the distribution,
$\qthetapost$, that is closest to $\pthetapost\left(\theta\right)$ in
Kullback-Liebler (KL) divergence within a restricted class $\mathcal{Q}$. We
consider the case where $\mathcal{Q}$ is a class of products of exponential
family distributions \citep{bishop:2006:pattern}:
\begin{eqnarray}\label{eq:kl_minimization}
\qthetapost &:=&
  \textrm{argmin}_{q \in \mathcal{Q}} \left\{KL(q || p)\right\}
  \quad \textrm{for} \nonumber  \\
\mathcal{Q} &=& \left\{q: q(\theta)  = \prod_{k=1} q(\theta_k);
  q(\theta_k) \propto e^{\eta_k ^T \theta_k}, \forall k \right\}
\nonumber\\
  KL &:=& -\left(\mbeq\left[ \log p\left(x \vert \theta \right)\right] +
           \mbeq\left[ \log p\left(\theta \vert \alpha \right) \right]\right) + \nonumber \\
    &&              \mbeq\left[ \log q \left(\theta \right) \right] + \textrm{Constant}.
\end{eqnarray}
We assume that $\qthetapost$, the solution to \eq{kl_minimization}, has interior
exponential family parameter $\eta_k$.  In this case, $\qthetapost$ can be
completely characterized by its mean parameters, $m := \expectq{\theta}$
\citep{wainwright2008graphical}.

It is well-known that optimal $\qthetapost$ chosen from this particular
$\mathcal{Q}$ under-estimates the posterior variance of $\theta$ (and provides
no estimate of the covariance between distinct $\theta_k$) even when the
posterior means are well-estimated \citep{turner:2011:two,
wang:2005:inadequacy}.  In order to improve the posterior covariance estimates,
\citet{giordano:2015:lrvb} estimates derivatives of the posterior cumulant
generating function by perturbing the objective in \eq{kl_minimization}, giving
the LRVB covariance estimate\footnote{See \citet{giordano:2015:lrvb} for
discussion of how to efficiently calculate and invert $\frac{\partial^2
KL}{\partial m \partial m^T}$.}
\begin{eqnarray}\label{eq:lrvb_covariance}
\hat\Sigma &:=& \left(\frac{\partial^2 KL}{\partial m \partial m^T} \right)^{-1}.
\end{eqnarray}
The LRVB approximation \eq{lrvb_covariance} is exactly equal to
$\Sigma := \textrm{Cov}_p(\theta)$ when VB estimates the
posterior means exactly. If the posterior mean estimates from VB are close to
the truth, then $\hat{\Sigma}$ can form a good approximation of $\hat\Sigma$.
In our experiments described in section \mysec{microcredit} we find this to be
the case. In \mysec{robustness}, we will discuss how the idea of LRVB can be
extended to provide local robustness measures.

\section{Measuring robustness with LRVB}\label{sec:robustness}

A typical end product of a Bayesian analysis might be a posterior expectation of
some function $\gtheta$ (e.g., a mean or variance): $\epgtheta$, which is a
functional of $g$. We suppose that we have determined that the prior parameter
$\alpha$ belongs to some set $\mathcal{A}$, perhaps after expert prior
elicitation. Finding the extrema of $\epgtheta$ as $\alpha$ ranges over all of
$\mathcal{A}$ is intractable or difficult except in special cases
\citep{moreno:2012:globalrobustness}. An alternative is to examine how much
$\epgtheta$ changes locally in response to small perturbations in the value of
$\alpha$:
\begin{eqnarray}\label{eq:local_robustness}
\left. \frac{d\epgtheta}{d\alpha} \right|_{\alpha} \Delta \alpha.
\end{eqnarray}
That is, we consider \emph{local robustness}
properties in lieu of global ones \citep{gustafson:2012:localrobustnessbook}.
By calculating \eq{local_robustness} for all
$\Delta \alpha \in \mathcal{A} - \alpha$, we can estimate the robustness
of $\epgtheta$ in a small neighborhood of $\alpha$.  For the rest of the paper
we will take $g(\theta) = \theta$ for simplicity.

Appendix A of \citet{giordano:2015:lrvb} shows that the derivation of the
approximate covariance given in \eq{lrvb_covariance}
arise as a special case of a general perturbation formula,
\begin{eqnarray}\label{eq:perturbed_elbo}
q_t &:=& \textrm{argmin}_{q \in \mathcal{Q}} \left\{KL + f(m)^T t \right\} \nonumber \\
\left. \frac{d\mathbb{E}_{q_t}\left[\theta\right]}{dt^T} \right|_{t=0} &=&
  \left(\frac{\partial^2 KL}{\partial m \partial m^T} \right)^{-1}
    \frac{\partial f(m)}{\partial m}.
\end{eqnarray}
By choosing an appropriate $f(m)$ in \eq{perturbed_elbo}, we can calculate estimates
of the local prior sensitivity from \eq{local_robustness}.
Let $\alpha_t := \alpha + \delta_\alpha t$ be the prior parameter perturbed
in the direction $\delta_\alpha$ by a small scalar amount $t$, so that
$\delta_\alpha t$ plays the role of $\Delta \alpha$ in \eq{local_robustness}.
For example, we could measure the sensitivity
of $\mbe_q[\theta]$ to the $i^{th}$ component of $\alpha$ by taking
$\delta_\alpha$ to be a vector of all zeros except with a $1$ in the $i^{th}$ place.

To tidy up notation, define
$\ell(\alpha, m) := \mbe_q\left[\log p(\theta \vert \alpha)\right]$.
Note that $\ell(\alpha, m)$ is a smooth function of $m$, since it is an
expectation with respect to the exponential family $q$, which is completely
parameterized by $m$.
If we additionally assume that $\log p(\theta \vert \alpha)$
is a smooth function of $\alpha$, then by a Taylor expansion in $\delta_\alpha t$,
\begin{eqnarray*}
\ell\left(\alpha + t \delta_\alpha, m\right) &=&
 \ell\left(\alpha, m\right) + \frac{\partial \ell }{\partial \alpha^T} \delta_\alpha t + O(t^2).
\end{eqnarray*}
We can then estimate the sensitivity of $\mbe_q[\theta]$ to the change $\delta_\alpha t$
in the prior parameter:
\begin{eqnarray}\label{eq:finite_dim_perturbation}
q_t &:=& \textrm{argmin}_{q \in \mathcal{Q}}
  \left\{KL + \frac{\partial \ell }{\partial \alpha^T} \delta_\alpha t \right\}
  \nonumber \\
\left. \frac{d\mathbb{E}_{q_t}\left[\theta\right]}{dt} \right|_{t=0} &=&
  \left(\frac{\partial^2 KL}{\partial m \partial m^T} \right)^{-1}
  \frac{\partial^2 \ell}{\partial m \partial \alpha^T}
    \delta_\alpha.
\end{eqnarray}
This easy-to-calculate closed-form expression is the LRVB approximation to the
local prior sensitivity \eq{local_robustness} to the change $\delta_\alpha t$.
As in \citet{giordano:2015:lrvb}, these derivatives are in fact the exact
sensitivity of the variational posterior expectations to prior perturbation.
The extent to which it represents the true prior sensitivity depends on the
extent to which the VB means are good estimates of the true posterior means.

\section{Microcredit experiment}\label{sec:microcredit}

\newcommand{\mcPriorEta}{15.01}
\newcommand{\mcPriorMuInfoMu}{0.03}
\newcommand{\mcPriorMuInfoTau}{0.02}
\newcommand{\mcPriorScaleAlpha}{20.01}
\newcommand{\mcPriorScaleBeta}{20.01}
\newcommand{\mcPriorTauAlpha}{2.01}
\newcommand{\mcPriorTauBeta}{2.01}

We apply the methods of \mysec{basic_lrvb} and \mysec{robustness} to a
hierarchical model from \citet{meager:2015:microcredit}. Randomized controlled
trials were run in seven different sites to try to measure the effect of access
to microcredit on various measures of business success, household poverty
indicators, and community welfare. However, it was unclear what if any
generalizable information had been learned about microcredit that could be
applied to other settings for policy purposes.
Thus, \citet{meager:2015:microcredit} fit a series of Bayesian hierarchical
models to estimate the general impact of microcredit on poor households and
assess the heterogeneity in this impact across the studies. For the purposes of
demonstrating robust Bayes techniques with VB, we will focus on the simpler of
the two models in \citet{meager:2015:microcredit} and ignore covariate
information.

We will index sites with $k=1,\dots,K$ (here, $K=7$) and businesses within a
site by $n=1,\dots,N_k$ ($N_k$ ranged from 961 to 16560).  In site $k$ and
business $n$ we observe whether the business was randomly selected for increased
access to microcredit, denoted $T_{nk}$, and the profit after intervention,
$y_{nk}$.  We follow \citet{rubin:1981:estimation} and assume that each site has
an idiosyncratic average profit, $\mu_k$, and average improvement in profit,
$\tau_k$, due to the intervention. Given $\mu_k$, $\tau_k$, and $T_{nk}$, the
observed profit is assumed to be generated with variance $\sigma_k^2$ according
to
\begin{eqnarray*}
y_{nk} \vert \mu_k, \tau_k, T_{nk}, \sigma_{k} &\indep&
  N\left(\mu_k + T_{nk} \tau_k, \sigma^2_{k} \right).
\end{eqnarray*}
The site effects, $(\mu_k, \tau_k)$, are assumed to be drawn independently
from an overall pool of effects.  For a given $k$, $\mu_k$ and $\tau_k$
may be correlated.
\begin{eqnarray}\label{eq:mu_model}
\left( \begin{array}{c} \mu_k \\ \tau_k \end{array}\right) \indep
N\left(
  \left( \begin{array}{c} \mu \\ \tau \end{array}\right), C \right)
\end{eqnarray}
The effects $\mu$, $\tau$, $\sigma_k^2$, and the covariance matrix $C$ are
unknown parameters that require priors.
For the covariance matrix $C$, we followed the recommended
practice of the software package Stan \citep{stan-manual:2015}
and used the non-conjugate LKJ prior
\cite{lewandowski:2009:lkj} with covariance parameter
$\eta = \mcPriorEta$ and inverse scale prior
$\Gamma(\mcPriorScaleAlpha, \mcPriorScaleBeta)$.
We used a conjugate gamma
prior on $\sigma_k^{-2} \iid \Gamma(\mcPriorTauAlpha, \mcPriorTauBeta)$.
Finally, for $(\mu, \tau)$ we used a bivariate normal prior:
\begin{eqnarray}\label{eq:mu_prior}
\left( \begin{array}{c} \mu \\ \tau \end{array}\right) \sim
N\left(
  \left( \begin{array}{c} \mu_0 \\ \tau_0 \end{array}\right), \Lambda^{-1} \right).
\end{eqnarray}
We used $\mu_0 = \tau_0 = 0$, and $\Lambda$ with entries $\mcPriorMuInfoMu$
and $\mcPriorMuInfoTau$ on the diagonals and zero off-diagonal.

To generate the MCMC samples, we used Stan \citep{stan-manual:2015}.  To
calculate all the derivatives and Hessians necessary for VB and LRVB we
implemented the objective function in C++ and then used the autodifferentiation
library of Stan \citep{carpenter:2015:stanautodiff}.

\subsection{Speed and validity of LRVB}

VB was over an order of magnitude faster than MCMC. Generating one set of 2500 MCMC
draws took 45 minutes. Optimizing the VB objective and calculating the LRVB
estimates, including all the reported sensitivity measures, took 58 seconds.

As can be seen in \fig{PosteriorGraphsMeans}, the VB and
MCMC posterior means are nearly identical, indicating that the necessary
assumptions for LRVB hold.  Next,
\fig{PosteriorGraphsSD} shows that the ordinary VB standard
deviations underestimate the true posterior standard deviations (as measured by
MCMC, which we take to be the ground truth), but that LRVB provides a good
correction.

\begin{knitrout}
\definecolor{shadecolor}{rgb}{0.969, 0.969, 0.969}\color{fgcolor}\begin{figure}[ht!]

{\centering \includegraphics[width=0.98\linewidth,height=0.4\linewidth]{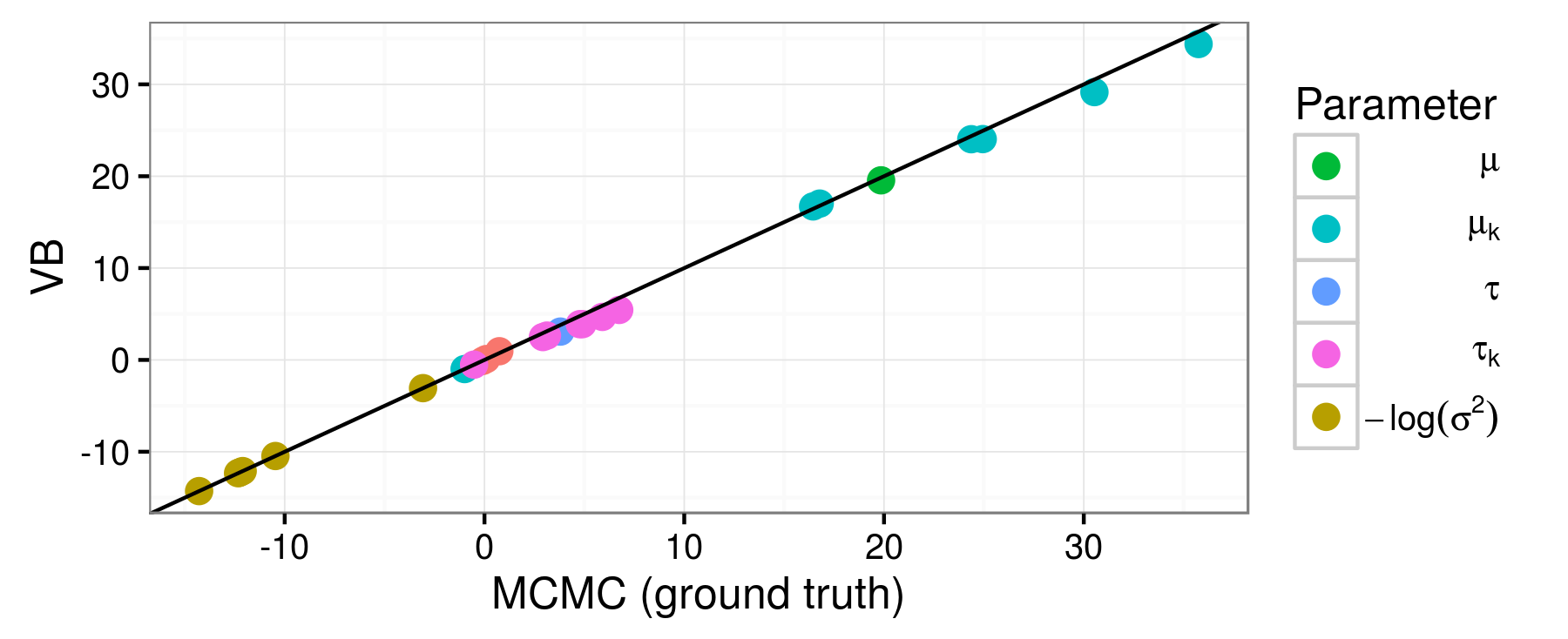} 

}

\caption[Posterior mean comparison]{Posterior mean comparison.}\label{fig:PosteriorGraphsMeans}
\end{figure}

\end{knitrout}

\begin{knitrout}
\definecolor{shadecolor}{rgb}{0.969, 0.969, 0.969}\color{fgcolor}\begin{figure}[ht!]

{\centering \includegraphics[width=0.98\linewidth,height=0.4\linewidth]{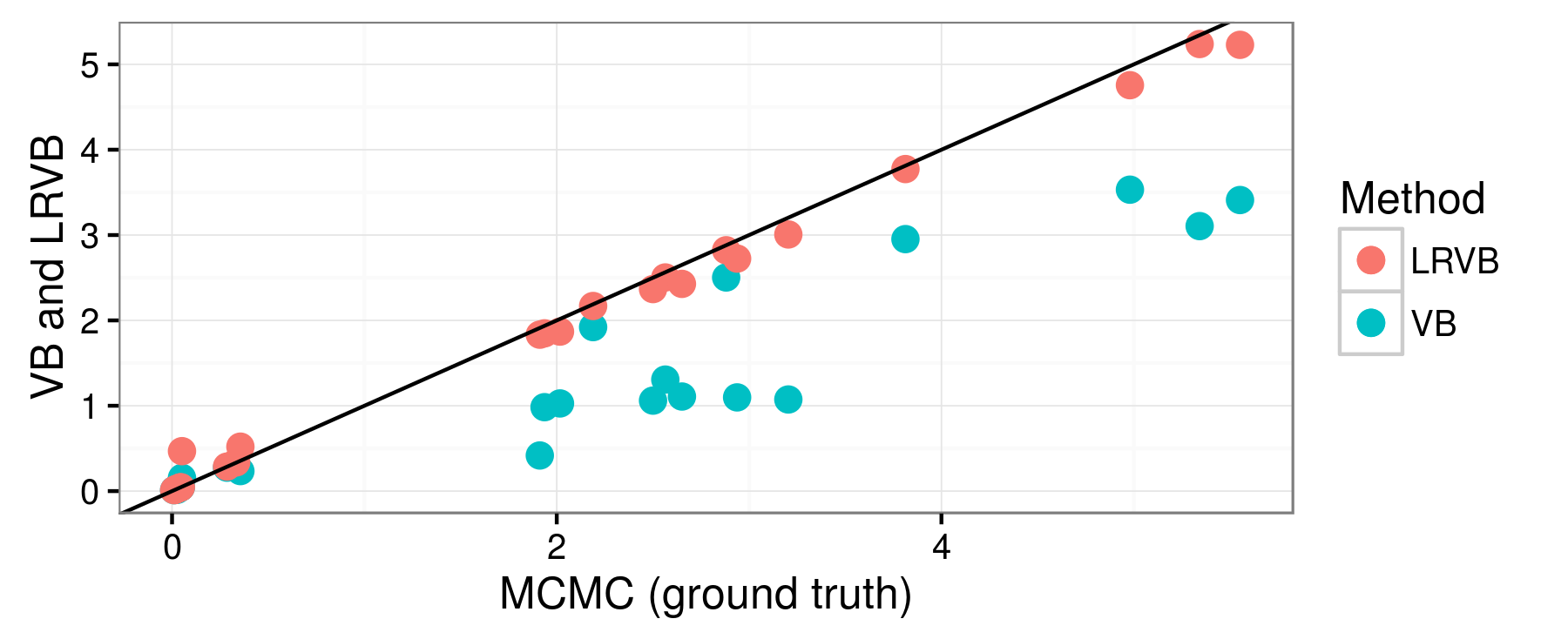} 

}

\caption[Posterior standard deviation comparison]{Posterior standard deviation comparison.}\label{fig:PosteriorGraphsSD}
\end{figure}

\end{knitrout}

Next, we turn to the evaluation of robustness, with an emphasis on the prior on
$(\mu, \tau)$. In \fig{InfluenceGraphsManual} we compare our LRVB robustness
estimates to the (extremely time-consuming) effect of manually changing a prior
parameter and re-running the MCMC chain.  Specifically, we changed
$\Lambda_{11}$ from $0.03$ to $0.04$ and measured how the change in the
posterior mean compared with the change predicted by LRVB.  The results
in \fig{InfluenceGraphsManual} show that the LRVB sensitivity estimates
match the actual sensitivity very closely.  Since VB estimates the
means reasonably well, as shown in \fig{PosteriorGraphsMeans}, and
\eq{finite_dim_perturbation} gives the exact sensitivity of the VB means to
prior perturbations, \fig{InfluenceGraphsManual} should not come as a surprise,
but it is a reassuring sanity check.

\begin{knitrout}
\definecolor{shadecolor}{rgb}{0.969, 0.969, 0.969}\color{fgcolor}\begin{figure}[ht!]

{\centering \includegraphics[width=0.98\linewidth,height=0.4\linewidth]{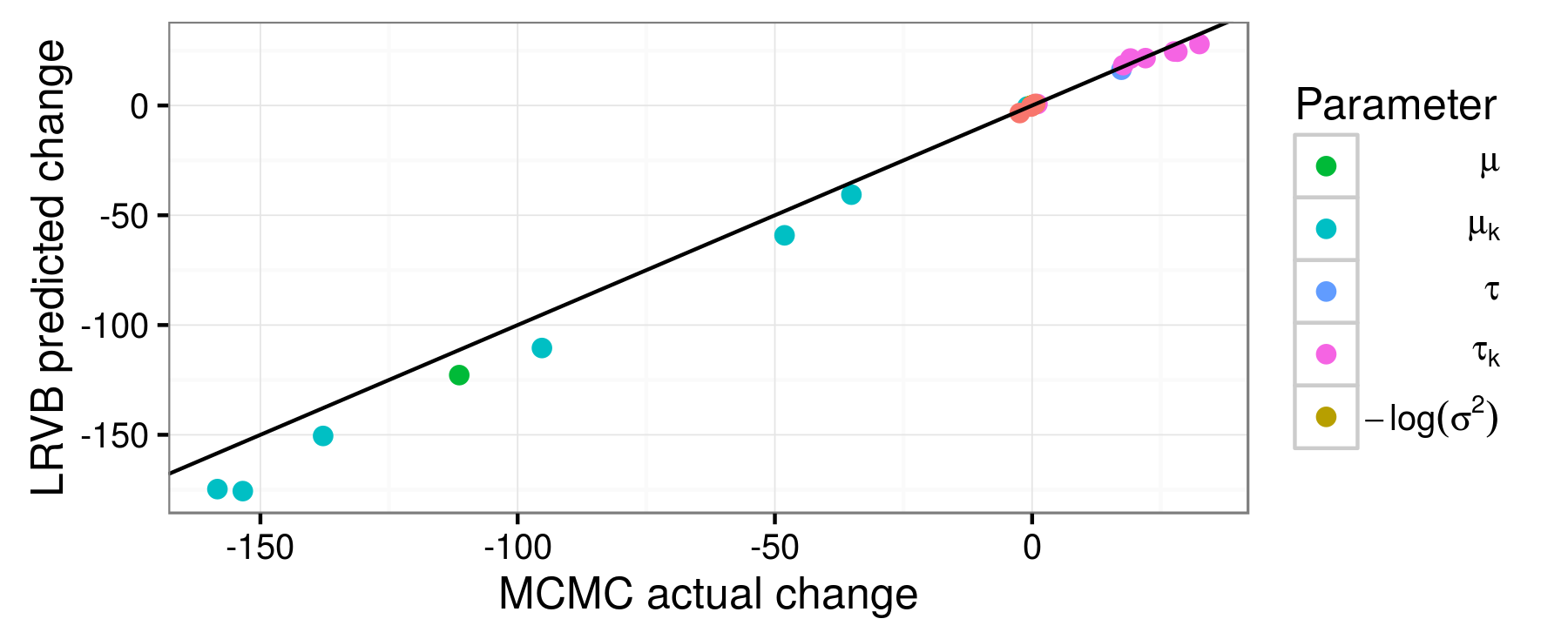} 

}

\caption[Manually perturbing and re-running MCMC]{Manually perturbing and re-running MCMC.}\label{fig:InfluenceGraphsManual}
\end{figure}

\end{knitrout}

\subsection{Analysis results}

\newcommand{\TauMean}{3.08}
\newcommand{\TauSd}{1.83}
\newcommand{\TauNumSds}{1.68}
\newcommand{\TauSens}{8.88}

Finally, we examine what our results tell us about microcredit within the
context of this particular simple model and prior choice.\footnote{We caution
that we are currently elaborating on methodology, not attempting to
take a stand on the value of microcredit.  Such a stand would require more
extensive and sophisticated analysis.  See \citet{meager:2015:microcredit} for
more extended discussion and modeling.}  We will focus on the parameter $\tau$
in \eq{mu_model}, which is intended to represent the overall ``global'' average
microcredit effectiveness.

The VB posterior mean and LRVB standard deviation for $\tau$ are
\begin{eqnarray*}
\mbeq\left[ \tau \right] = \TauMean \quad \quad \quad
\textrm{StdDev}_{q}\left( \tau \right) = \TauSd.
\end{eqnarray*}
Under a normal assumption on the posterior of $\tau$, this does not provide
strong evidence for the effectiveness of microcredit, since the mean is
only $\TauNumSds$ standard deviations from zero.

However, the standard deviation is not necessarily the full story. We might also
ask whether, if our priors were different, we might have come to a different
conclusion. The sensitivity of $\tau$ to its prior, \eq{mu_prior}, is shown in
\fig{InfluenceGraphsTau}.  In this graph, we report the sensitivity in units of
posterior standard deviations of $\tau$ in order to show how one can affect
posterior inference by changing the prior parameter. In particular, notice that
$\mbeq\left[\tau\right]$ is quite sensitive to the prior parameter $\Lambda$.
For example, the first bar in \fig{InfluenceGraphsTau} shows that increasing
$\Lambda_{11}$ by $0.04$ would increase $\mbeq\left[\tau\right]$ by $0.04 \cdot
\TauSens$ standard deviations, which would be enough to make $\tau$ look
significantly greater than zero.  $\mbeq\left[\tau\right]$ is even more
sensitive to $\Lambda_{12} (=\Lambda_{21})$, the off-diagonal covariance terms.
As seen in the second graph of \fig{InfluenceGraphsTau}, it is not particularly
sensitive to its prior mean.

The meaning of non-robustness results like this depends on the modeler's
beliefs about the prior.  In this case, the question is whether we think
that a change of $0.04$ in $\Lambda_{11}$ or other similar influental
perturbations indicated by \fig{InfluenceGraphsTau} would be reasonable
expressions of prior uncertainty.  This decision must always depend on the
context.  If a reasonable range of priors could lead to a range of posterior
means that greatly exceeds the spread of the original posterior, then
the posterior standard deviation must represent an under-estimate of
subjective uncertainty.  In any case, as a prerequisite to making such decisions,
the modeler needs to be able to measure the robustness, and this measurement
is made easily available through LRVB.

\begin{knitrout}
\definecolor{shadecolor}{rgb}{0.969, 0.969, 0.969}\color{fgcolor}\begin{figure}[ht!]

{\centering \includegraphics[width=0.98\linewidth,height=0.4\linewidth]{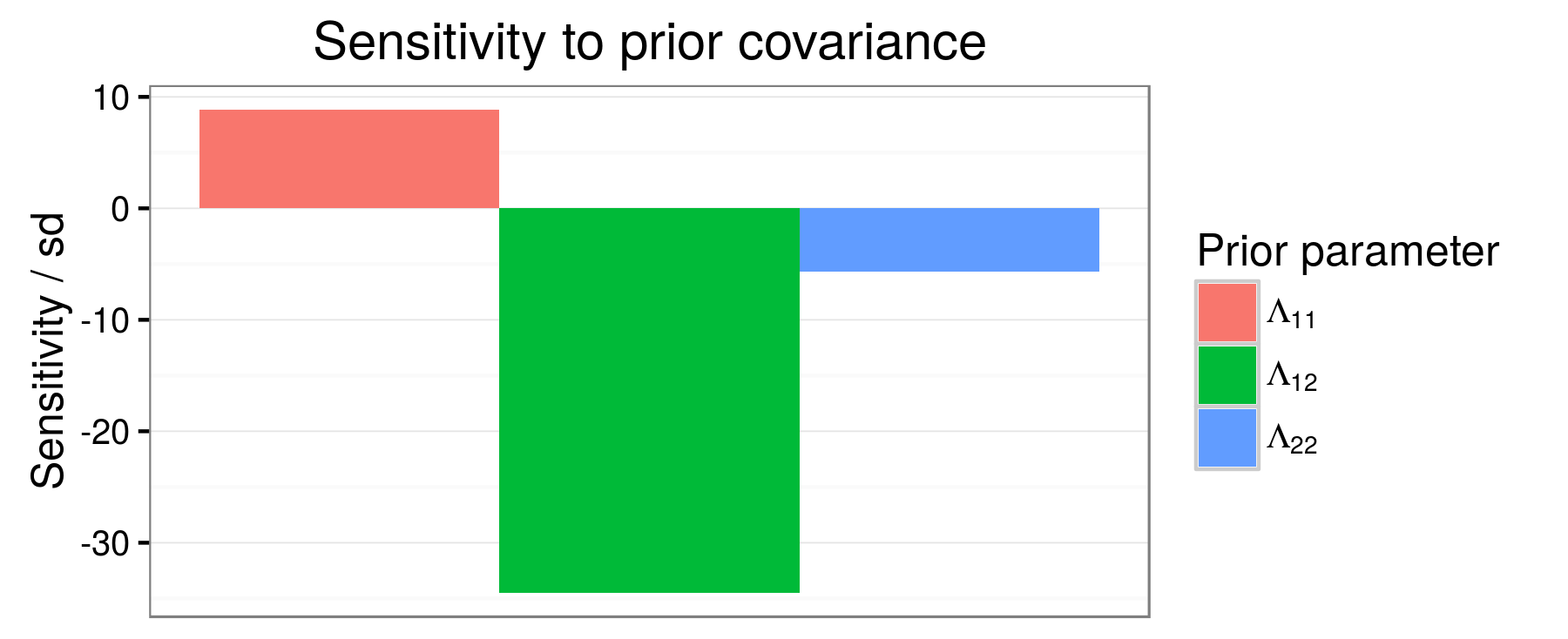} 
\includegraphics[width=0.98\linewidth,height=0.4\linewidth]{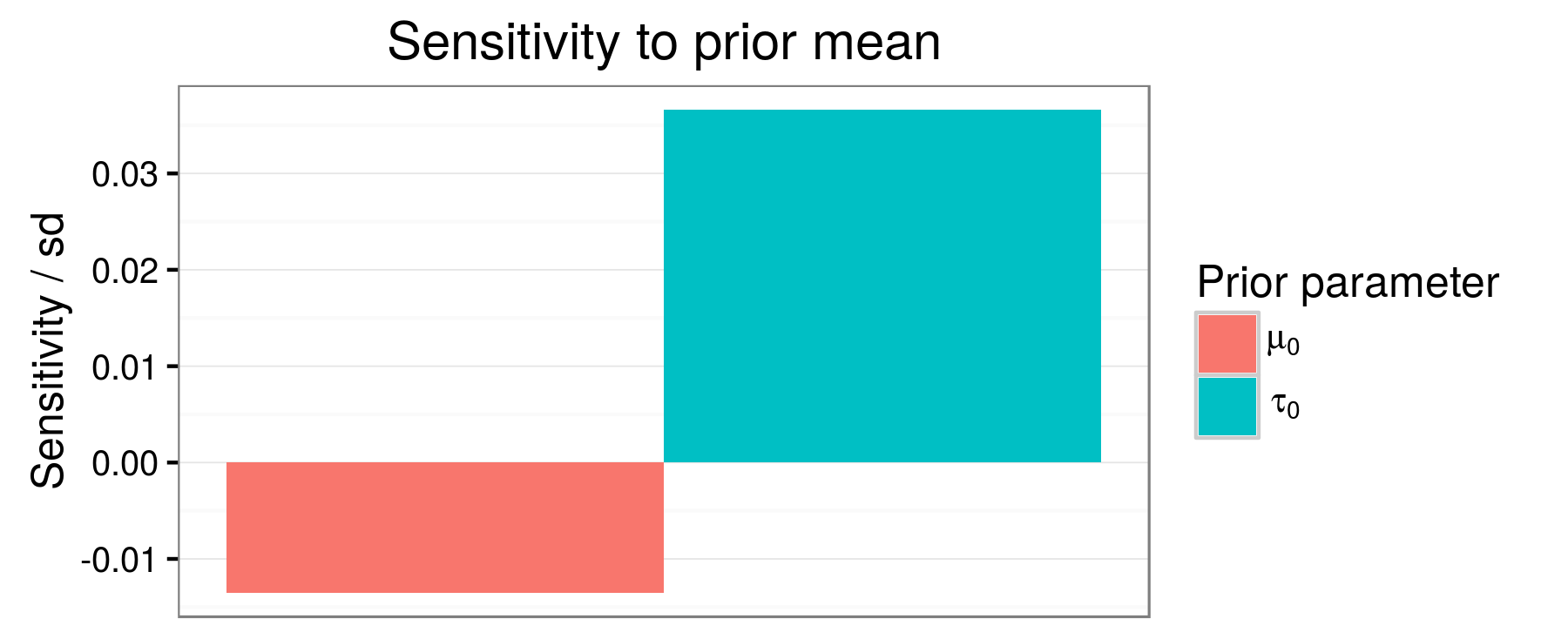} 

}

\caption[Normalized sensitivity of tau]{Normalized sensitivity of tau}\label{fig:InfluenceGraphsTau}
\end{figure}

\end{knitrout}

\section{Conclusion}

Hierarchical models are a valuable tool for the social sciences, but
they can be slow to fit with MCMC.  Furthermore, they can suffer from
non-robustness in the form of sensitivity to the choice of priors,
and MCMC does not provide an easy-to-use, general-purpose robustness measure.

VB, together with LRVB, can provide good approximations to Bayesian posteriors
over an order of magnitude faster than MCMC.  Furthermore, LRVB also
provides easy-to-calculate measures of robustness that can alert the
modeler to excessive prior sensitivity.


{\small
\bibliographystyle{icml2016}
\bibliography{influence_scores}
}

\end{document}